\documentclass{article}

\usepackage[utf8]{inputenc}
\usepackage{amsmath, amssymb}
\usepackage{graphicx}
\usepackage{natbib}
\usepackage{booktabs}
\usepackage{hyperref}
\hypersetup{
    colorlinks=true,
    linkcolor=blue,
    filecolor=magenta,      
    urlcolor=cyan,
    pdftitle={A Stable Loss Function for Direct Preference Optimization},
    pdfpagemode=FullScreen,
    }

\title{A Principled Loss Function for Direct Language Model Alignment}
\author{
  Yuandong Tan\\\\
  \texttt{kyoani@stu.pku.edu.cn }
}

\begin{document}

\maketitle

\begin{abstract}
The alignment of large language models (LLMs) with human preferences is commonly achieved through Reinforcement Learning from Human Feedback (RLHF). Direct Preference Optimization (DPO) simplified this paradigm by establishing a direct mapping between the optimal policy and a reward function, eliminating the need for an explicit reward model. However, we argue that the DPO loss function is theoretically misaligned with its own derivation, as it promotes the indefinite maximization of a logits difference, which can lead to training instability and reward hacking. In this paper, we propose a novel loss function derived directly from the RLHF optimality condition. Our proposed loss targets a specific, finite value for the logits difference, which is dictated by the underlying reward, rather than its maximization. We provide a theoretical analysis, including a gradient-based comparison, to demonstrate that our method avoids the large gradients that plague DPO when the probability of dispreferred responses approaches zero. This inherent stability prevents reward hacking and leads to more effective alignment. 
\end{abstract}

\section{Introduction}

Aligning the behavior of large language models (LLMs) with human values and preferences is a critical step in their development and deployment. Reinforcement Learning from Human Feedback (RLHF) has emerged as the de facto standard for this task \citep{ouyang2022training}. The goal of RLHF is to optimize a language model policy, $\pi_\theta$, to align with human preferences. This is typically framed as maximizing a reward function, $r(x,y)$, learned from a preference dataset $\mathcal{D}_p = \{(x, y_w, y_l)\}$, where $y_w$ is preferred over $y_l$ for a prompt $x$. The standard RLHF objective is:
\begin{equation}
\max_{\pi_\theta} \mathbb{E}_{x\sim\mathcal{D}, y\sim\pi_{\theta}} [r(x,y) ] - \beta \text{KL}[\pi_{\theta}(y|x) \Vert\pi_{\text{ref}}(y|x)],
\label{eq:RLHF}
\end{equation}
where the KL-divergence term, scaled by $\beta$, regularizes the policy to stay close to a reference model $\pi_\text{ref}$.

\citet{rafailov2023direct} showed that this objective can be optimized more directly with Direct Preference Optimization (DPO). They established a theoretical link between the optimal policy $\pi^*$ and the underlying reward function:
\begin{equation}
r(x,y) = \beta \log \frac{\pi^*(y|x)}{\pi_\text{ref}(y|x)} + Z(x),
\label{eq:DPO_reward}
\end{equation}
where $Z(x)$ is a normalization term dependent only on $x$. This insight allows DPO to bypass the explicit training of a reward model.

However, we identify a fundamental contradiction in the DPO objective. While its theoretical derivation points to a specific relationship between the optimal policy and the reward, its loss function promotes the unbounded maximization of the log-probability ratio between the preferred and dispreferred responses. This maximization objective is only consistent with the theory in the edge case where the reward difference between responses is infinite. For any finite reward, this objective is misaligned. Furthermore, this objective can lead to pathologically large gradients, especially when the probability of the dispreferred response ($\pi_\theta(y_l|x)$) becomes very small, a phenomenon that can cause training instability and reward hacking.

To address these shortcomings, we introduce a new loss function. Our approach stems directly from the optimality condition of Equation \ref{eq:RLHF}. Instead of maximizing the log-probability ratio, our loss function optimizes the policy towards a specific target value for this ratio, which is determined by the reward difference. This principled objective leads to a more stable optimization landscape. Specifically, our loss function naturally dampens gradients when the logits difference is large, preventing the instability seen in DPO and mitigating the risk of reward hacking.

\section{Methodology}

\subsection{From Optimality Condition to a New Objective}
From the relationship in Equation \ref{eq:DPO_reward}, we can express the difference in rewards between a winning ($y_w$) and losing ($y_l$) completion for a given prompt $x$ under the optimal policy $\pi^*$ as:
\begin{equation}
r(x, y_w) - r(x, y_l) = \beta \left( \log \frac{\pi^*(y_w|x)}{\pi_\text{ref}(y_w|x)} - \log \frac{\pi^*(y_l|x)}{\pi_\text{ref}(y_l|x)} \right).
\end{equation}
This equation defines the condition for an optimal policy. The goal of training should be to steer the current policy $\pi_\theta$ to satisfy this condition. Let us define the policy-dependent logits difference as:
\begin{equation}
\text{logits}(\pi_\theta) = \log \frac{\pi_\theta(y_w|x)}{\pi_\text{ref}(y_w|x)} - \log \frac{\pi_\theta(y_l|x)}{\pi_\text{ref}(y_l|x)}.
\end{equation}
The optimality condition is therefore met when:
\begin{equation}
\text{logits}(\pi_\theta) = \frac{r(x, y_w) - r(x, y_l)}{\beta}.
\label{eq:optimality}
\end{equation}
This reveals that the optimal policy does not require maximizing the logits, but rather driving them towards a finite target value. The DPO loss, which is equivalent to a log-sigmoid loss on $\beta \cdot \text{logits}(\pi_\theta)$, encourages making $\text{logits}(\pi_\theta)$ infinitely large, contradicting the theoretical foundation.

\subsection{The Proposed Stable Preference Loss}
We require a loss function that reaches its minimum when the optimality condition in Equation \ref{eq:optimality} is met. Althoughugh a squared error loss, $(\text{logits} - \frac{r_w - r_l}{\beta})^2$, 

We propose a more robust loss function whose structure inherently guides the logits to a stable point. Consider the function $f(z) = -z e^{-z}$, which has a unique global maximum at $z=1$. We can leverage this property. Let $z = c \cdot \text{logits}$, where $c$ is a scaling constant. We want the loss to be minimized when $\text{logits} = 1/c$. From Equation \ref{eq:optimality}, this implies $c \approx \frac{\beta}{r_w - r_l}$.

By absorbing the unknown reward difference into the hyperparameter $\beta$, we formulate our loss of stable preference optimization (SPO) as follows:
\begin{equation}
\mathcal{L}_{\text{SPO}}(\theta) = -(\beta \cdot \text{logits}(\pi_\theta)) \exp(-\beta \cdot \text{logits}(\pi_\theta)) - \left(\alpha \cdot \log \frac{\pi_{\text{ref}}(y_l|x)}{\pi_\theta(y_l|x)}\right) \exp\left(-\alpha \cdot \log \frac{\pi_{\text{ref}}(y_l|x)}{\pi_\theta(y_l|x)}\right)
\end{equation}
This loss is minimized when its argument, $\beta \cdot \text{logits}(\pi_\theta)$equals 1. This provides a clear and stable optimization target: $\text{logits}(\pi_\theta) = 1/\beta$. This formulation has several advantages:
\begin{enumerate}
    \item \textbf{Principled Target:} It optimizes towards a finite, stable point consistent with RLHF theory.
    \item \textbf{Robustness to Over-Optimization:} As $\text{logits}(\pi_\theta) \to \infty$, the loss gracefully decays to zero. This prevents the model from being penalized for being "too confident," avoiding unstable gradients for well-distinguished pairs.
    \item \textbf{Asymmetric Penalty:} The loss function heavily penalizes logits values less than the target $1/\beta$, while applying a vanishing penalty for values greater than the target.
    \item \textbf{Stable Preference Learning:} While it is difficult to precisely estimate the magnitude of $\frac{\pi_\theta(y_w|x)}{\pi_{\text{ref}}(y_w|x)}$, we know that as long as $\frac{\pi_\theta(y_l|x)}{\pi_{\text{ref}}(y_l|x)}$decreases and the difference between the two ratios remains within a certain range, the model will effectively learn human preferences. This approach prevents simultaneous sharp declines in the probabilities of both positive and negative examples, and avoids overfitting on positive examples. 
\end{enumerate}

\section{Theoretical Analysis: Comparison with DPO}
\subsection{Gradient Analysis}
The core advantage of our SPO loss becomes evident through a gradient analysis. Let $\Pi_w = \pi_\theta(y_w|x)$ and $\Pi_l = \pi_\theta(y_l|x)$. The DPO loss is $\mathcal{L}_{\text{DPO}} = -\log \sigma(\beta \cdot \text{logits})$. Its gradient with respect to the policy probability of the losing response, $\Pi_l$, is:
\begin{equation}
\frac{\partial \mathcal{L}_{\text{DPO}}}{\partial \pi_\theta(y_l|x)} = \beta \cdot \sigma(-\beta \cdot \text{logits}) \cdot \frac{1}{\pi_\theta(y_l|x)}.
\end{equation}
As the model learns, it will often drive $\Pi_l \to 0$ for clear preference pairs. In this scenario, $\text{logits} \to \infty$. While the term $\sigma(-\beta \cdot \text{logits})$ approaches zero, the term $1/\Pi_l$ approaches infinity. The product can lead to extremely large gradients, causing training instability. This large gradient on $\Pi_l$ can backpropagate to create a large negative gradient on $\Pi_w$ as well, forcing the model to reduce the probability of \textit{both} responses to escape the punitive gradient signal. This is a form of reward hacking.

Now, consider the gradient of our proposed SPO loss. Let $X = \beta \cdot \text{logits}$. The derivative of $\mathcal{L}_{\text{SPO}}$ w.r.t $X$ is $\frac{\partial \mathcal{L}_{\text{SPO}}}{\partial X} = (X-1)e^{-X}$. The gradient w.r.t $\Pi_l$ is:
\begin{equation}
\frac{\partial \mathcal{L}_{\text{SPO}}}{\partial \Pi_l} = \frac{\partial \mathcal{L}_{\text{SPO}}}{\partial X} \frac{\partial X}{\partial \Pi_l} = -(\beta \cdot \text{logits} - 1)e^{-\beta \cdot \text{logits}} \cdot \frac{\beta}{\Pi_l}.
\end{equation}
Crucially, as $\Pi_l \to 0$ and $\text{logits} \to \infty$, the exponential term $e^{-\beta \cdot \text{logits}}$ decays to zero much faster than $1/\Pi_l$ grows. This powerful exponential decay dominates the expression, forcing the entire gradient to zero. This property ensures that once the model has learned a preference pair sufficiently well (i.e., the logits difference is large), the gradient vanishes, leading to a stable and robust training process that is immune to this specific form of gradient explosion and reward hacking.

\section{Experiments}
\subsection{Experimental Setup}
To validate the effectiveness of our proposed SPO loss, we conduct a comprehensive set of experiments on two leading base models: \textbf{Qwen2.5-7B-Instruct} and \textbf{Llama-3-8B-Instruct}. Our fine-tuning process consists of two stages:
\begin{enumerate}
    \item \textbf{Supervised Fine-Tuning (SFT)}: We first fine-tune the base models on the \texttt{HuggingFaceH4/ultrachat\_200k}~\citep{ultrachat_200k} dataset to enhance their general instruction-following capabilities. This results in our SFT baseline models.
    \item \textbf{Preference Alignment}: Following SFT, the models are further aligned using preference data. We compare our SPO method against the standard DPO baseline using the \texttt{HuggingFaceH4/ultrafeedback\_binarized}~\citep{ultrafeedback_binarized} dataset. Both the SPO and DPO alignment runs start from the same SFT checkpoint for a fair comparison.
\end{enumerate}
We evaluated the final models by conducting pairwise, head-to-head comparisons and using GPT-4 as the judge to determine a win rate. We report the win rates for all three model versions (SFT, DPO, and our SPO) against each other.

\subsection{Results}
The experimental results, presented in Table \ref{tab:qwen_results} and Table \ref{tab:llama_results}, unequivocally demonstrate the superiority of our proposed SPO loss. In the tables, each cell shows the win rate of the model in the row against the model in the column.

For the \textbf{Qwen2.5-7B} model (Table \ref{tab:qwen_results}), we observe a clear hierarchy of performance. The DPO model vastly outperforms the SFT baseline with a \textbf{91.70\%} win rate. More importantly, our SPO model achieves a significant improvement over DPO, securing a \textbf{56.50\%} win rate in a direct head-to-head matchup. Against the SFT baseline, our SPO model's superiority is even more pronounced, with a staggering \textbf{95.15\%} win rate.

\begin{table}[h!]
\centering
\caption{Win rates for \textbf{Qwen2.5-7B} fine-tuning methods. Each cell shows the win percentage of the row model against the column model.}
\label{tab:qwen_results}
\begin{tabular}{@{}lccc@{}}
\toprule
\textbf{Win Rate (\%) of Row vs. Column} & \textbf{SFT} & \textbf{DPO} & \textbf{SPO (Ours)} \\
\midrule
\textbf{SFT}         & --   & 8.30 & 4.85 \\
\textbf{DPO}         & 91.70 & --   & 43.50 \\
\textbf{SPO (Ours)}  & \textbf{95.15} & \textbf{56.50} & -- \\
\bottomrule
\end{tabular}
\end{table}

We confirmed this trend by repeating the experiment on the \textbf{Llama-3-8B} model, with the results shown in Table \ref{tab:llama_results}. The pattern of improvement is remarkably consistent. DPO shows a strong gain over SFT (91.46\% win rate). Once again, our SPO method delivers a clear performance boost over DPO, winning the head-to-head comparison with a \textbf{53.73\%} win rate.

\begin{table}[h!]
\centering
\caption{Win rates for \textbf{Llama-3-8B} fine-tuning methods. Each cell shows the win percentage of the row model against the column model.}
\label{tab:llama_results}
\begin{tabular}{@{}lccc@{}}
\toprule
\textbf{Win Rate (\%) of Row vs. Column} & \textbf{SFT} & \textbf{DPO} & \textbf{SPO (Ours)} \\
\midrule
\textbf{SFT}         & --   & 8.54 & 6.32 \\
\textbf{DPO}         & 91.46 & --   & 46.27 \\
\textbf{SPO (Ours)}  & \textbf{93.68} & \textbf{53.73} & -- \\
\bottomrule
\end{tabular}
\end{table}

Across both model architectures, the results are unambiguous: SFT provides a solid foundation, DPO offers a substantial improvement through preference alignment, and our SPO method consistently and significantly outperforms DPO. This validates that the benefits of SPO's stable and principled loss function generalize across different models, leading to a more effective and robust alignment with human preferences.

\section{Conclusion}
In this paper, we introduced a novel loss function, Stable Preference Optimization (SPO), for aligning language models with human preferences. We identified a theoretical inconsistency in the widely-used DPO method, where the loss function promotes an unbounded maximization of logits, conflicting with the finite optimal value derived from RLHF theory. This can lead to training instability and reward hacking. Our proposed SPO loss resolves this issue by optimizing towards a specific, finite target for the logits difference, which is grounded in the theoretical optimality condition. A gradient analysis confirms that our method is robust to the gradient explosion that can affect DPO. Empirical results validate our theory, showing that our method significantly outperforms DPO and produces models that are competitive with larger, state-of-the-art models. We believe that SPO provides a more stable, principled, and effective path for future research in language model alignment.

\bibliographystyle{plainnat}
\bibliography{references}


\end{document}